%% file: main.tex
\documentclass[10pt,reqno]{amsart}
\usepackage[T1]{fontenc}
\usepackage{lmodern}
\usepackage[utf8]{inputenc}
\usepackage[letterpaper,textwidth=6in,textheight=8in,centering]{geometry}

\usepackage{amsmath,amssymb,mathtools,bm}
\usepackage{booktabs,array}
\usepackage{enumitem}
\usepackage{microtype}
\usepackage{graphicx}
\usepackage{xcolor}
\usepackage{caption}
\usepackage{placeins}
\usepackage{needspace}
\usepackage{makecell}
\usepackage[numbers,sort&compress]{natbib}
\usepackage{xurl}
\usepackage{hyperref}

\makeatletter
\renewcommand\paragraph{\@startsection{paragraph}{4}{\z@}%
  {0.8ex plus 0.2ex minus 0.1ex}{-0.5em}{\normalfont\bfseries}}
\makeatother

\theoremstyle{plain}
\newtheorem{theorem}{Theorem}[section]
\newtheorem{proposition}[theorem]{Proposition}

\newtheorem{corollary}[theorem]{Corollary}
\theoremstyle{definition}

\theoremstyle{remark}
\newtheorem{remark}[theorem]{Remark}

\graphicspath{{figs/}}

\hypersetup{
  hidelinks,
  pdftitle={Late-Time Dynamics of Flow Matching: Distance Geometry and Learning},
  pdfauthor={Umberto Biccari, Junqing Huang, Kang Liu, and Enrique Zuazua}
}
\setlist{nosep}

\newcommand{\R}{\mathbb{R}}
\newcommand{\E}{\mathbb{E}}
\newcommand{\Y}{\mathcal{Y}}

\newcommand{\dd}{\mathrm{d}}
\newcommand{\eps}{\varepsilon}

\newcommand{\dist}{\operatorname{dist}}
\newcommand{\conv}{\operatorname{conv}}

\newcommand{\Pc}{\mathcal P}
\newcommand{\partialC}{\partial^{\mathrm C}}
\newcommand{\Law}{\operatorname{Law}}

\numberwithin{equation}{section}

\DisableLigatures{encoding = *, family = * }

\title[Late-Time Dynamics of Flow Matching]{Late-Time Dynamics of Flow Matching: Distance Geometry and Learning}

\author[U. Biccari]{Umberto Biccari}
\address{Chair of Computational Mathematics, DeustoTech, University of Deusto,
Avenida de las Universidades 24, 48007 Bilbao, Basque Country, Spain}
\email{umberto.biccari@deusto.es}

\author[J. Huang]{Junqing Huang}
\address{Chair of Computational Mathematics, DeustoTech, University of Deusto,
Avenida de las Universidades 24, 48007 Bilbao, Basque Country, Spain}
\email{junqing.huang@deusto.es}

\author[K. Liu]{Kang Liu}
\address{Institut de Math\'ematiques de Bourgogne, Universit\'e de Bourgogne, CNRS,
9 Avenue Alain Savary, 21000 Dijon, France}
\email{kang.liu@u-bourgogne.fr}

\author[E. Zuazua]{Enrique Zuazua}
\address{Chair for Dynamics, Control, Machine Learning, and Numerics,
Department of Mathematics, Friedrich-Alexander-Universit\"at Erlangen-N\"urnberg, Germany;
Chair of Computational Mathematics, DeustoTech, University of Deusto, Spain;
Departamento de Matem\'aticas, Universidad Aut\'onoma de Madrid, Spain}
\email{enrique.zuazua@fau.de}

\date{November 20, 2026}

\begin{document}

\begin{abstract}
Generative models transform simple random inputs into samples that resemble observed data. For a finite training set, a central question is how its geometry shapes the terminal dynamics and the probabilities assigned to individual samples. We address this question for flow matching with linear interpolation between independent Gaussian and empirical samples. The squared distance to the training set generates explicit paths contracting exponentially toward the nearest sample. We prove that exact trajectories restarted at prescribed late-time states approximate these paths uniformly over all elapsed times and compact sets of starting states away from Voronoi interfaces, with an error exponentially small in the inverse squared restart noise scale. The initial mass near interfaces controls the quadratic Wasserstein error between the exact and reference laws. For Gaussian and fixed Gaussian-tube restart laws, this gives a square-root rate in the restart noise scale, uniform over time, which a two-site example shows to be sharp in general. At ties, the Clarke subdifferential characterizes the graphical limit of the rescaled velocity field. This geometry motivates two learning methods: fitting the distance potential with value and gradient supervision, and fitting velocities along explicit contraction paths. A transport estimate separates initialization mismatch, velocity error, incomplete contraction, and endpoint allocation. MNIST experiments illustrate the distinct roles of support proximity, nearest-site retention, and feature fidelity. These results connect flow matching and score-based diffusion through a common limiting geometry and show how asymptotic analysis can inform the design and evaluation of generative models.
\end{abstract}

\maketitle

\section{Introduction}\label{sec:introduction}

A dynamical generative model starts with samples from a simple probability
distribution and moves them toward a distribution represented by data.
When the data consist of finitely many points, this task has a geometric
aspect: trajectories must approach those points. It also has a
probabilistic aspect: the points must receive the appropriate proportions
of samples. Understanding how these two aspects interact can guide both
the analysis of generative dynamics and the construction of their
learning targets.

Flow matching learns a velocity field from prescribed paths connecting
source and data samples. For linear interpolation, each training pair
traces a straight segment, although trajectories of the corresponding
marginal velocity can curve \citep{lipman2023flow,liu2023rectified}.
Score-based diffusion provides another dynamical description: Gaussian
noise smooths the data distribution, and the score, its logarithmic
gradient, defines a probability flow toward decreasing noise
\citep{ho2020ddpm,song2019score,song2021sde,karras2022edm}. For independent
Gaussian and empirical samples, an exact change of space and time relates
these two flows \citep[Propositions~2.1-2.2]{wan2025denoisers}. The final
stage of flow matching can therefore be studied through the geometry of
Gaussian-smoothed finite data as the noise vanishes.

The relevant geometric object is the squared distance to the training
set $\Y$. Away from points with several nearest neighbors, its scaled
negative gradient points toward the unique nearest training sample.
Starting from $x$, the resulting motion has the explicit form
\[
 F_s(x)=P_\Y(x)+e^{-s/2}\bigl(x-P_\Y(x)\bigr),
 \qquad s\geq0,
\]
where $P_\Y(x)$ denotes that nearest sample. The path remains in its
initial Voronoi cell and contracts exponentially toward its site. Thus a
single nearest-neighbor query determines an entire reference trajectory.
At the cell boundaries, the Clarke subdifferential describes the possible
limiting velocities.

Our main result quantifies how this reference motion approximates exact
flow matching after a prescribed late restart. If the exact dynamics
start at $x$ at logarithmic time $\ell$, Theorem~\ref{thm:main}(ii) bounds
their distance from $F_s(x)$ by
\[
 C_K\exp(-c_Ke^\ell),
\]
uniformly over every elapsed time $s\geq0$ and every $x$ in a compact set
$K$ separated from Voronoi interfaces. Here $e^\ell$ is the inverse
squared noise scale at restart. This approximation concerns the entire
remaining path. The theorem also distinguishes prescribed restarts from
time translates of a single convergent trajectory, whose limiting paths
are constant. Corollary~\ref{cor:fm-endpoints} supplies individual
endpoint convergence and rates.

For an ensemble of starting positions, the interfaces play a different
role. A small displacement of a boundary between endpoint basins may send
nearby initial states to different training points. The trajectory error
in cell interiors can therefore be much smaller than the transport error
between the output laws. Corollary~\ref{cor:fm-macroscopic} gives an
all-time Wasserstein approximation governed by the source mass near these
interfaces. Gaussian and fixed Gaussian-tube restart laws yield the rate
$O(e^{-\ell/4})$, or the square root of the restart noise scale. A
two-site calculation shows that this exponent is sharp in general.

The explicit geometry suggests two learning constructions.
\emph{Potential-Distance Learning} (PDL) fits the squared-distance
potential together with the spatial gradient used during generation.
\emph{Potential-Distance Flow Matching} (PDFM) fits the velocities along
the reference contractions. Their training targets require nearest-site
queries and explicit formulas; sampling the PDFM supervision paths
requires no numerical integration. Generation then evolves the learned
potential-gradient or velocity field. Under a one-sided expansion bound,
Proposition~\ref{prop:weighted-transport} controls the output law using
the velocity residual along reference paths, incomplete contraction at
finite time, and the discrepancy between source cell masses and target
weights. Remark~\ref{rem:other-initial-laws} includes a change of source
law. This decomposition links the training objective to transport
accuracy while making the role of initialization explicit.

\paragraph{Relation to earlier analysis and learning methods.}
\citet{wan2025denoisers} derive projection limits and terminal convergence
rates for Gaussian flow matching with discrete and manifold-supported
data. Their Propositions~5.9-5.10 establish late-time nearest-site
absorption and endpoint convergence for exact and asymptotically accurate
denoisers. The estimate here controls the entire remaining trajectory,
uniformly over compact sets of prescribed starting states. The complete
Clarke graph follows from the heat-score geometry of
\citet{liuzuazua2026convergence} through the Gaussian correspondence.
The assignment cells of the full flow studied by
\citet{pierret2026tessellations} may differ from the Voronoi cells of the
reference contractions. Prescribed or semidiscrete couplings in
conditional flow matching can enforce target probabilities
\citep{pooladian2023multisample,tong2024otcfm,mousavihosseini2026semidiscrete}.
For the reference dynamics considered here, each site receives the
initial probability of its Voronoi cell.

Data geometry also enters projection interpretations of diffusion
\citep{permenter2024projection}, energy-based constructions
\citep{balcerak2025energy,wang2025equilibrium}, and distance-field
learning. Distance Marching \citep{wang2026distance} studies generation
from distance fields and the influence of nearest-neighbor assignments
and initialization. Our analysis derives squared-distance supervision
from finite-support asymptotics and gives a transport estimate for the
resulting learned paths. Continuous normalizing flows and neural ODEs
\citep{chen2018node,grathwohl2019ffjord}, including semiautonomous neural
ODE approximation \citep{li2026sanode}, provide the broader setting of
learned transport dynamics. The jumps of the nearest-site velocity across
interfaces make density-weighted approximation especially natural here.

The numerical study compares PDL and PDFM with denoising score matching,
heat-noise prediction, and flow matching on MNIST. A common noisy
empirical initialization and a common heat-coordinate sampling interval
allow the five methods to be assessed through support proximity,
occupancy, and held-out feature statistics. PDL retains the initial
nearest-site assignment most frequently, while FM attains the lowest
feature-distribution discrepancies. The two-site example illustrates
how interface motion changes the endpoint law.

\paragraph{Organization of the paper.}
Section~\ref{sec:foundations} introduces the exact generative equations
and their correspondence. Section~\ref{sec:theory} states the late-time
results, and Section~\ref{sec:method} develops PDL, PDFM, and transport
control. Section~\ref{sec:proofs} contains the proofs, followed by the
numerical study in Section~\ref{sec:numerical-experiments} and the
conclusions in Section~\ref{sec:conclusion}. Network architectures and evaluation details are provided in
Appendix~\ref{supp:implementation}, and generation sequences in
Appendix~\ref{supp:images}.

\section{Exact generative dynamics and their correspondence}
\label{sec:foundations}
Let $\Y=\{y_1,\ldots,y_N\}\subset\R^d$ contain distinct training points,
and let $\mu_N=N^{-1}\sum_i\delta_{y_i}$. We write $\Pc_2(\R^d)$ for
probability measures with finite second moment and $W_2$ for their quadratic
Wasserstein distance. A density and its probability measure use the same
symbol. Throughout the paper, \emph{flow matching} refers to linear
interpolation between independent $D\sim\mu_N$ and
$G\sim\gamma_d:=\mathcal N(0,I_d)$. The independence assumption specifies
the coupling used in the analysis; other couplings give different marginal
fields \citep{pooladian2023multisample,tong2024otcfm}.
We use $t$ for each model's native time and $\tau$ for logarithmic
generation time. This section concerns exact fields; benchmark training
objectives are given in Section~\ref{sec:numerical-experiments}.

\subsection{Heat-score probability flow}
\label{sec:foundation-score}\label{sec:pde-forward-reverse}
We begin with the heat equation initialized at the empirical data measure:
\begin{equation*}
 \begin{cases}
 \partial_tu=\Delta u,&x\in\R^d,\ t>0,\\
 u(\cdot,0)=\mu_N.
 \end{cases}
\end{equation*}

The measure-valued initial condition is understood by weak convergence as
$t\to0^+$. Its solution is the Gaussian convolution
\begin{equation*}
 u_t(x)=(G_t*\mu_N)(x)=\frac1N\sum_{i=1}^N G_t(x-y_i),
 \qquad G_t(z)=\frac{e^{-|z|^2/(4t)}}{(4\pi t)^{d/2}}.
\end{equation*}

Equivalently, $u_t$ is the law of $D+\sqrt{2t}\,G$: heat evolution
replaces each training point by a Gaussian cloud of covariance $2tI_d$.
For positive time this density is smooth and strictly positive, so its
logarithmic spatial gradient, or \emph{score}, is well defined:
\begin{equation*}
 s(x,t)=\nabla\log u_t(x)=\frac{\bar y_t^S(x)-x}{2t},\qquad
 \bar y_t^S(x)=\frac{\sum_i y_iG_t(x-y_i)}{\sum_jG_t(x-y_j)}.
\end{equation*}

The Gaussian-weighted mean $\bar y_t^S(x)$ averages the training sites
according to their kernel values at $x$. Since $s(\cdot,t)u_t=\nabla u_t$, the same
marginal evolution satisfies the continuity equation
\begin{equation*}\label{eq:heat-continuity}
 \partial_tu_t+\nabla\!\cdot(-s(\cdot,t)u_t)=0.
\end{equation*}

Its characteristics give deterministic generation when followed toward
decreasing heat time \citep{song2021sde,liuzuazua2025pde}. For $T>0$ and
$x_T\in\R^d$, the heat probability-flow equation is
\begin{equation}\label{eq:det-reverse-original-time}
 \frac{\dd X_t^S}{\dd t}=-s(X_t^S,t),\qquad
 X_T^S=x_T,\qquad 0<t\leq T.
\end{equation}

The score is spatially Lipschitz with linear growth on each positive-time
strip, and the equation has a unique solution on $(0,T]$. With random
initial state $X_T^S\sim u_T$, its marginal at heat time $t$ is $u_t$.
The finite-horizon source $u_T$ is a Gaussian mixture centered at the data.

\subsection{Flow matching}\label{sec:foundation-cnf-fm}
Consider the linear interpolation
\begin{equation}\label{eq:straight-fm-path}
 C_t=(1-t)G+tD,\qquad m_t=\Law(C_t),\qquad 0\leq t\leq1.
\end{equation}

Each pair traces a straight segment with velocity $D-G$. The marginal
velocity averages these pair velocities at the current state
\citep{lipman2023flow,liu2023rectified}.

\begin{proposition}[Interpolation and velocity]\label{prop:fm-continuity}
The curve $m\in\operatorname{AC}^2([0,1];\Pc_2(\R^d))$ joins $\gamma_d$
to $\mu_N$. There exists $p\in C([0,1)\times\R^d;\R^d)$ such that
\begin{equation*}
 \partial_tm_t+\nabla_x\!\cdot(p_tm_t)=0
 \quad\text{in }\mathcal D'((0,1)\times\R^d).
\end{equation*}
For $t<1$, the positive smooth density and a corresponding velocity are
\begin{align}
 m_t(x)&=\frac1N\sum_iK_t(x,y_i),\qquad
 K_t(x,y)=G_{(1-t)^2/2}(x-ty), \notag 
 \\
 p_t(x)&=\frac{\bar y_t^F(x)-x}{1-t},\qquad
 \bar y_t^F(x)=\frac{\sum_i y_iK_t(x,y_i)}{\sum_jK_t(x,y_j)}.
 \label{eq:fm-marginal-velocity}
\end{align}
\end{proposition}
The proof in Section~\ref{sec:proof-interpolation} derives the velocity
by averaging the fluxes of the affine Gaussian components.

For $x_0\in\R^d$, the exact generative trajectory solves
\begin{equation}\label{eq:cnf-ode}
 \frac{\dd X_t^F}{\dd t}=p_t(X_t^F),\qquad
 X_0^F=x_0,\qquad 0\leq t<1.
\end{equation}

The velocity is spatially Lipschitz with linear growth on every strip
$0\leq t\leq b<1$, hence this solution is unique on $[0,1)$.
With $X_0^F\sim\gamma_d$, its marginal law is $m_t$.
The conditional sample paths in \eqref{eq:straight-fm-path} are straight,
whereas the integral curves of their averaged velocity are generally curved.

\subsection{Exact equivalence with heat probability flow}
\label{sec:exact-equivalence}
The Gaussian noise-to-signal transformation
\citep[Propositions~2.1-2.2]{wan2025denoisers} gives an exact space-time
correspondence between these deterministic equations. Introduce
\begin{equation*}
 h:(0,1)\longrightarrow(0,\infty),\qquad
 h(t)=\frac{(1-t)^2}{2t^2},\qquad
 h^{-1}(r)=\frac{1}{1+\sqrt{2r}}.
\end{equation*}
The map is a smooth decreasing bijection. The associated Gaussian
identities are
\begin{align}
 m_t(x)&=t^{-d}u_{h(t)}(x/t),\label{eq:fm-heat-density}\\
 p_t(x)&=\frac{x}{t}+\frac{1-t}{t^2}s(x/t,h(t)).
 \label{eq:exact-field-correspondence}
\end{align}

\begin{proposition}[Exact trajectory equivalence]\label{prop:exact-equivalence}
For any $x_0\in\R^d$, let $X^F$ solve \eqref{eq:cnf-ode}.
For $t_0\in(0,1)$, set $T=h(t_0)$, and let $X^S$ be the unique heat
probability-flow solution of~\eqref{eq:det-reverse-original-time} on $(0,T]$ with
$X_T^S=X_{t_0}^F/t_0$. Then
\begin{equation*}
 \begin{aligned}
 X_t^F&=tX_{h(t)}^S, &&t_0\leq t<1,\\
 X_r^S&=\frac{X_{h^{-1}(r)}^F}{h^{-1}(r)}, &&0<r\leq T.
 \end{aligned}
\end{equation*}
Conversely, let $T>0$ and let $X^S$ solve
\eqref{eq:det-reverse-original-time} with $X_T^S=x_T$.
With $t_0=h^{-1}(T)$, the first identity defines the unique flow-matching
trajectory on $[t_0,1)$ with $X_{t_0}^F=t_0x_T$; it extends uniquely to
$[0,1)$.
\end{proposition}
The proof is given in Section~\ref{sec:proof-equivalence}. The time $t_0$
marks the beginning of the matched heat interval; the flow-matching trajectory starts at zero.
Under Gaussian initialization, its matched heat state has law $u_{h(t_0)}$.
A finite heat horizon therefore corresponds to a terminal interval of flow
matching; the full Gaussian-source evolution corresponds to infinite heat
time. The equivalence includes a spatial dilation and concerns exact
probability-flow ODEs.

\section{The squared-distance limiting system}\label{sec:theory}
We ask what path an exact flow-matching trajectory follows when initialized at a prescribed state late in generation. The heat correspondence from Section~\ref{sec:exact-equivalence} identifies the limiting squared-distance dynamics, including the directions that arise at Voronoi interfaces. Away from those interfaces, we show in Theorem \ref{thm:main} that the exact trajectories approach explicit contractions toward their nearest sites, uniformly over compact sets of starting states. 

\paragraph{Nearest-point geometry.} The Voronoi cells of the empirical support \(\Y=\{y_1,\ldots,y_N\}\) determine the form of the limiting dynamics. Set
\[
I(x)=\operatorname*{arg\,min}_{1\le i\le N}|x-y_i|^2,
\qquad
\Sigma_N=\{x\in\mathbb R^d:|I(x)|>1\}.
\]
For $x\notin\Sigma_N$, denote its unique nearest training point by
\begin{equation*}
 P_\Y(x)=\operatorname*{arg\,min}_{y\in\Y}|x-y|^2.
\end{equation*}
The squared-distance potential and the associated velocity are
\[
\Phi_N(x)=\operatorname{dist}(x,\Y)^2=\min_i|x-y_i|^2,
\qquad
b_N(x)=\frac12\bigl(P_\Y(x)-x\bigr).
\]

On each open cell, \(\Phi_N\) is quadratic and \(b_N=-\nabla\Phi_N/4\). At a tie, the Clarke generalized gradient \cite{clarke1990optimization} is the convex hull of the gradients from the adjacent cells:
\[
\partial^C\Phi_N(x)=2\bigl(x-\operatorname{conv}\{y_i:i\in I(x)\}\bigr).
\]
The limiting dynamics therefore take the form
\begin{equation}\label{eq:limiting-inclusion}
 \dot Z_s\in-\frac14\partialC\Phi_N(Z_s),\qquad s\geq0.
\end{equation}

We interpret \eqref{eq:limiting-inclusion} almost everywhere along locally absolutely continuous paths. A point \(z\) is an equilibrium if \(0\in\partial^C\Phi_N(z)\).

For \(x\notin\Sigma_N\), the segment joining \(x\) to \(P_\Y(x)\) lies entirely in the same open cell. The solution from \(x\) thus follows a single affine vector field and is given by
\begin{equation}\label{eq:hard-flow-path}
 F_s(x)=P_\Y(x)+e^{-s/2}\bigl(x-P_\Y(x)\bigr).
\end{equation}

These explicit contractions are the reference paths in Theorem \ref{thm:main}(ii) and the supervision paths in Section \ref{sec:method}. For expectations under densities, we use any fixed nearest-site rule on \(\Sigma_N\), which has Lebesgue measure zero.

\paragraph{Logarithmic generation time.}
For the exact flow-matching trajectory, set
\begin{equation}\label{eq:fm-log-clock}
 t_F(\tau)=1-e^{-\tau/2},\qquad Y_\tau^F=X_{t_F(\tau)}^F,
 \qquad \dot Y_\tau^F=B_{t_F(\tau)}(Y_\tau^F),
\end{equation}
where
\begin{equation*}\label{eq:rescaled-fm-field}
 B_t(x)=\frac{1-t}{2}p_t(x)=\frac12\bigl(\bar y_t^F(x)-x\bigr),
\end{equation*}
and $\bar y_t^F(x)$ is defined in \eqref{eq:fm-marginal-velocity}.

Thus the terminal regime $t\to1^-$ becomes $\tau\to\infty$.
A time-shift limit is a subsequential limit of $Y^F_{\tau_n+\cdot}$,
$\tau_n\to\infty$, in $C_{\mathrm{loc}}([0,\infty);\R^d)$.
Part~(i) below considers translates of one exact trajectory; part~(ii)
restarts the dynamics from a prescribed state at logarithmic time $\ell$.

\Needspace{10\baselineskip}

\begin{theorem}[Late-time dynamics of flow matching]\label{thm:main}
The rescaled flow-matching dynamics \eqref{eq:fm-log-clock} satisfy:
\begin{enumerate}[label=(\roman*),leftmargin=*]
\item \textbf{Limiting subgradient dynamics.}
The rescaled fields converge graphically:
\[
 \operatorname{gph}B_t\longrightarrow
 \operatorname{gph}\!\left(-\frac14\partialC\Phi_N\right)
 \qquad(t\to1^-).
\]
The convergence to $b_N$ is locally uniform off $\Sigma_N$.
For every deterministic initial state and every $\tau_n\to\infty$,
there is a subsequence such that
\[
 Y^F_{\tau_{n_k}+\cdot}\longrightarrow Z
 \quad\text{in }C_{\mathrm{loc}}([0,\infty);\R^d).
\]
Every such limit solves \eqref{eq:limiting-inclusion}.

\item \textbf{Uniform convergence for all elapsed times.}
Fix a compact $K\subset\R^d\setminus\Sigma_N$. For $x\in K$, let
\[
 \dot Z_s^{\ell,x}=B_{t_F(\ell+s)}(Z_s^{\ell,x}),\qquad
 Z_0^{\ell,x}=x.
\]
There exist $C_K,c_K>0$ and $\ell_K\geq0$, depending on $K$ and
the training support, such that, for every $\ell\geq\ell_K$,
\begin{equation}\label{eq:finite-window-limit}
 \sup_{x\in K}\sup_{s\geq0}|Z_s^{\ell,x}-F_s(x)|
 \leq C_K\exp(-c_Ke^\ell),
\end{equation}
where $F_s$ is the explicit contraction in \eqref{eq:hard-flow-path}.
\end{enumerate}
\end{theorem}

At a tie $x$, graphical convergence characterizes the possible limits of
$B_t(x_t)$ as $t\to1^-$ and $x_t\to x$: they form exactly
$-\partialC\Phi_N(x)/4$.
The following corollaries give individual endpoint rates and an all-time
Wasserstein approximation for restarted ensembles. Time shifts of a fixed
convergent trajectory have constant limits, whereas prescribed late
restarts retain the contraction paths used for learning. Proofs are given
in Section~\ref{sec:proof-main}.

\begin{corollary}[Microscopic convergence: endpoints and rates]\label{cor:fm-endpoints}
Every exact flow-matching trajectory \eqref{eq:cnf-ode} converges as $t\to1^-$ to an
equilibrium of \eqref{eq:limiting-inclusion}. With $X_0^F\sim\gamma_d$,
its endpoint $y^{F,*}$ belongs to $\Y$ almost surely, has law $\mu_N$,
and almost surely satisfies
\begin{equation*}
 |X_t^F-y^{F,*}|=O(1-t),\qquad
 |Y_\tau^F-y^{F,*}|=O(e^{-\tau/2}).
\end{equation*}
The constants may depend on the initial state.
\end{corollary}

\begin{corollary}[Macroscopic convergence and rate]
\label{cor:fm-macroscopic}
Let $q\in\Pc_2(\R^d)$ satisfy $q(\Sigma_N)=0$, and write
$\mathcal Z_s^\ell(x)=Z_s^{\ell,x}$ as in Theorem~\ref{thm:main}(ii).
Then $(\mathcal Z_s^\ell)_\#q$ converges to $(F_s)_\#q$ in $W_2$,
uniformly for $s\geq0$, as $\ell\to\infty$.
For the standard Gaussian law, any fixed heat marginal $u_T$, or any
fixed Gaussian-tube law \eqref{eq:tube-density} with $t_{\min}>0$,
\[
 \sup_{s\geq0}W_2\big((\mathcal Z_s^\ell)_\#q,(F_s)_\#q\big)
 =O(e^{-\ell/4}).
\]
The exponent $1/4$ is worst-case sharp for Gaussian initial laws over finite supports.
\end{corollary}

More generally, the proof gives $O(e^{-\alpha\ell/4})$ whenever $q(\{x:\operatorname{dist}(x,\Sigma_N)\le r\})=O(r^\alpha)$ as $r\downarrow0$. The constants for a tube source can depend on its fixed lower heat cutoff. The exponent describes the cost of reallocating interface mass even when individual paths deep inside cells are exponentially accurate.

\section{Learning the limiting dynamics}\label{sec:method}\label{sec:learning-targets}
The limiting dynamics in Theorem \ref{thm:main} give rise to two learning objectives. PDL fits the squared-distance potential and uses its gradient for generation. PDFM fits the velocity along the explicit contraction paths in \eqref{eq:hard-flow-path}. In each case, the network provides a differentiable representation of terminal geometry for repeated evaluation during generation. Proposition \ref{prop:weighted-transport} below bounds the effect of the PDFM path loss on the generated law. We use
$\tau_f>0$ for the terminal generation time in the logarithmic clock.
In this section, generation starts from $q_{\mathrm{tube}}$; other initial
laws are treated in Remark~\ref{rem:other-initial-laws}.
Learned generative equations are considered on intervals where they are well posed.

\paragraph{Gaussian tube sampling.}
Fix $T>0$, set $t_{\min}=Te^{-\tau_f}$, and define
\begin{equation}\label{eq:tube-density}
 \pi_T(t)=\frac{\mathbf1_{[t_{\min},T]}(t)}{\tau_f t},\qquad
 q_{\mathrm{tube}}(x)=\int_{t_{\min}}^T u_t(x)\pi_T(t)\,\dd t.
\end{equation}

Sampling $Q=D+\sqrt{2t}\,G$ with independent
$t\sim\pi_T$, $D\sim\mu_N$, and $G\sim\gamma_d$ gives this density.
Since $t=Te^{-\tau}$, it is the average of exact heat-flow marginals over
$0\leq\tau\leq\tau_f$ and concentrates supervision in noisy neighborhoods
of the data. The joint state-time law is
\begin{equation*}
 \rho_{\mathrm{tube}}(\dd x,\dd t)=u_t(x)\pi_T(t)\,\dd x\,\dd t.
\end{equation*}

PDL uses tube samples as query states; PDFM uses them as initial states
for its analytic contraction paths.

\subsection{Potential-Distance Learning}\label{sec:pdl}
Let $\Phi_\theta$ be a differentiable scalar network.
Since $q_{\mathrm{tube}}$ has a density and the tie set
$\Sigma_N$ is Lebesgue-null, the projection $P_\Y(Q)$
is uniquely defined almost surely for $Q\sim q_{\mathrm{tube}}$.
On $\Sigma_N$, we fix the measurable selection
$P_\Y(x)=y_{\min I(x)}$, which exists since $\Y$ is finite.
This convention does not affect the loss, and
$\nabla\Phi_N(Q)=2(Q-P_\Y(Q))$ almost surely.

For $\beta>0$, PDL minimizes
\begin{equation}\label{eq:pdl-augmented-loss}
 \mathcal L_{\mathrm{PDL}}^\beta(\theta)
 =\E_{Q\sim q_{\mathrm{tube}}}
 \Big[
 |\Phi_\theta(Q)-\Phi_N(Q)|^2
 +\beta|\nabla\Phi_\theta(Q)-2(Q-P_\Y(Q))|^2
 \Big].
\end{equation}
A single nearest-point query evaluates both targets. Generation follows
\begin{equation*}
 \dot Z_\tau=-\frac14\nabla\Phi_\theta(Z_\tau),\qquad
 Z_0\sim q_{\mathrm{tube}},\qquad 0\leq\tau\leq\tau_f.
\end{equation*}
The learned potential decreases along a classical trajectory, with
$\dd\Phi_\theta(Z_\tau)/\dd\tau=-|\nabla\Phi_\theta(Z_\tau)|^2/4$.

The two targets in \eqref{eq:pdl-augmented-loss} couple the geometry
of the potential to the velocity used during generation.

\subsection{Potential-Distance Flow Matching}\label{sec:pdfm}
With $q_{\mathrm{tube}}$ given by \eqref{eq:tube-density},
draw $X_0\sim q_{\mathrm{tube}}$ and set
$Y^*=P_\Y(X_0)$. As in Section~\ref{sec:pdl},
the nearest-point projection is unique almost surely,
and the same measurable selection applies on $\Sigma_N$.
The limiting system provides the path and velocity
\begin{equation}\label{eq:exp-pdfm-path}
 X_\tau=Y^*+e^{-\tau/2}(X_0-Y^*),\qquad
 V_\tau^*=\frac12(Y^*-X_\tau),\qquad 0\leq\tau\leq\tau_f.
\end{equation}

These are the paths \(X_\tau=F_\tau(X_0)\) from \eqref{eq:hard-flow-path}. A single nearest-site query at \(X_0\) determines both the path and its velocity, which PDFM fits by minimizing
\begin{equation}\label{eq:exp-pdfm-loss}
 \mathcal L_{\mathrm{PDFM}}(\theta)
 =\E_{X_0\sim q_{\mathrm{tube}}}
   \int_0^{\tau_f}|V_\theta(X_\tau,\tau)-V_\tau^*|^2\,\dd\tau.
\end{equation}

The field may be autonomous, $V_\theta(x)$, or time dependent,
$V_\theta(x,\tau)$; both representations approximate the same autonomous
target $b_N$. Generation solves
\begin{equation}\label{eq:pdfm-generation}
 \dot Z_\tau=V_\theta(Z_\tau,\tau),\qquad
 Z_0\sim q_{\mathrm{tube}},\qquad 0\leq\tau\leq\tau_f.
\end{equation}

The sampled paths induce the state law $(F_\tau)_\#q_{\mathrm{tube}}$ at time $\tau$. The loss measures velocity accuracy under these time-dependent distributions.

\subsection{Transport accuracy}\label{sec:weighting-rationale}\label{sec:inexactness}
Let $\mathcal V_i=\{x:|x-y_i|<|x-y_j|\ \text{for all }j\ne i\}$.
The reference contraction has endpoint law
\begin{equation*}
 \nu_*=(P_\Y)_\#q_{\mathrm{tube}}
      =\sum_i q_{\mathrm{tube}}(\mathcal V_i)\delta_{y_i}.
\end{equation*}

The loss in \eqref{eq:exp-pdfm-loss} measures velocity error along the reference paths. The next estimate bounds its effect on the output law and separates it from incomplete contraction and the discrepancy between the Voronoi cell masses and the empirical weights.

\Needspace{9\baselineskip}
\begin{proposition}[Path loss and transport error]\label{prop:weighted-transport}
Assume $V_\theta$ is continuous, locally Lipschitz in space uniformly on
compact time--space sets, and, for some $\lambda\in\R$, satisfies
\begin{equation}\label{eq:one-sided-bound}
 \langle V_\theta(x,\tau)-V_\theta(y,\tau),x-y\rangle
 \leq\lambda|x-y|^2
 \quad(x,y\in\R^d,\ 0\leq\tau\leq\tau_f).
\end{equation}
For $Z_0\sim q_{\mathrm{tube}}$, let $\nu_{\theta,\tau_f}$ be the
output law of \eqref{eq:pdfm-generation}. Then
\begin{equation}\label{eq:error-separation}
 W_2(\nu_{\theta,\tau_f},\mu_N)
 \leq\sqrt{\kappa_\lambda(\tau_f)\mathcal L_{\mathrm{PDFM}}(\theta)}
 +e^{-\tau_f/2}D_{\mathrm{tube}}+W_2(\nu_*,\mu_N),
\end{equation}
where
\[
 D_{\mathrm{tube}}=\|\dist(\cdot,\Y)\|_{L^2(q_{\mathrm{tube}})},
 \qquad \kappa_\lambda(a)=\int_0^a e^{2\lambda s}\,\dd s.
\]

\end{proposition}

The reference path retains its initial nearest site, giving the identity
\begin{equation}\label{eq:hard-flow-wasserstein-identity}
 W_2\big((F_\tau)_\#q_{\mathrm{tube}},\nu_*\big)
 =e^{-\tau/2}D_{\mathrm{tube}},\qquad \tau\geq0.
\end{equation}
For $N\ge2$, let $\Delta=\tfrac12\sum_i|q_{\mathrm{tube}}(\mathcal V_i)-1/N|$ and $m=\min_{i\ne j}|y_i-y_j|$. Every coupling of $\nu_*$ and $\mu_N$ moves at least $\Delta$ between distinct sites. Matching the common masses first moves exactly $\Delta$, so
\[
 m^2\Delta\le W_2^2(\nu_*,\mu_N)\le \operatorname{diam}(\Y)^2\Delta.
\]
Thus allocation depends on the labeled site masses. Two assignment histograms can have the same occupancy entropy and different transport costs. The learned field's one-sided expansion
constant $\lambda$ controls the amplification of pathwise velocity errors;
a spatial Lipschitz constant $L$ permits $\lambda=L$. Within a strict
Voronoi cell, the exact field contracts at rate $1/2$. Global expansion
also accounts for interactions across cells: for an autonomous learned
field with distinct equilibria $a,b$, \eqref{eq:one-sided-bound} gives
$0\leq\lambda|a-b|^2$, hence $\lambda\geq0$.

For PDL, the same coupling argument applies to
$V_\theta=-\nabla\Phi_\theta/4$, with path residual
\begin{equation}\label{eq:pdl-path-residual}
 \mathcal R_{\mathrm{PDL}}(\theta)
 =\frac1{16}\int_0^{\tau_f}\!\int_{\R^d}
 |\nabla\Phi_\theta(x)-\nabla\Phi_N(x)|^2
 \,\dd\big((F_s)_\#q_{\mathrm{tube}}\big)(x)\,\dd s.
\end{equation}

Under the regularity and one-sided hypotheses of Proposition~\ref{prop:weighted-transport}, the PDL transport estimate has the form
\eqref{eq:error-separation}, with $\mathcal L_{\mathrm{PDFM}}$ replaced
by $\mathcal R_{\mathrm{PDL}}$.
The training objective \eqref{eq:pdl-augmented-loss} weights derivatives
by the tube-query law; \eqref{eq:pdl-path-residual} weights them by the
reference-path occupation measure. This identifies the derivative
accuracy relevant to transport and provides a direct evaluation diagnostic
\citep{otto2001geometry,AGS08}.

\begin{remark}[Other initial laws]\label{rem:other-initial-laws}
If generation starts from $\nu_0\in\Pc_2(\R^d)$, denote the output law by
$\widetilde\nu_{\theta,\tau_f}$. Under the assumptions of
Proposition~\ref{prop:weighted-transport},
\begin{equation}\label{eq:initialization-mismatch}
 W_2(\widetilde\nu_{\theta,\tau_f},\nu_{\theta,\tau_f})
 \leq e^{\lambda\tau_f}W_2(\nu_0,q_{\mathrm{tube}}).
\end{equation}

This follows by transporting an optimal initial coupling through the
learned flow, whose trajectories satisfy
$|Z_\tau^x-Z_\tau^y|\leq e^{\lambda\tau}|x-y|$.
The right-hand side of \eqref{eq:initialization-mismatch} adds to
\eqref{eq:error-separation} for the changed initial law.

The same proof also gives a direct estimate using any source density
$q\in\Pc_2(\R^d)$. Define
\[
 \mathcal L_q(\theta)=\E_{X_0\sim q}\int_0^{\tau_f}
 |V_\theta(F_s(X_0),s)-b_N(F_s(X_0))|^2\,\dd s,
\]
$D_q=\|\dist(\cdot,\Y)\|_{L^2(q)}$, and $\nu_*^q=(P_\Y)_\#q$.
For the learned output $\nu_{\theta,\tau_f}^{q}$ from $q$,
\[
 W_2(\nu_{\theta,\tau_f}^{q},\mu_N)
 \leq\sqrt{\kappa_\lambda(\tau_f)\mathcal L_q(\theta)}
       +e^{-\tau_f/2}D_q+W_2(\nu_*^q,\mu_N).
\]

Taking $q=u_T$ evaluates the path residual under the numerical generation
source; the training loss in \eqref{eq:exp-pdfm-loss} uses
$q=q_{\mathrm{tube}}$.
\end{remark}

\section{Proofs of the analytical results}\label{sec:proofs}
We first prove the interpolation and trajectory correspondence, then establish
the limiting theorem and its two corollaries. The microscopic rate follows
from the heat-score endpoint analysis; the macroscopic rate combines
cellwise contraction with a quantitative interface-layer estimate.
The final subsection proves the path-weighted transport estimate.

\subsection{Gaussian interpolation}\label{sec:proof-interpolation}
\begin{proof}[Proof of Proposition~\ref{prop:fm-continuity}]
\leavevmode\par\smallskip\noindent
\textbf{Step 1: The interpolation law and its regularity.}
Write $g(z)=(2\pi)^{-d/2}e^{-|z|^2/2}$ and
$F_t^i(z)=(1-t)z+ty_i$. Independence and the empirical target law give
\begin{equation}\label{eq:app-fm-mixture}
 m_t=\frac1N\sum_{i=1}^N(F_t^i)_\#\gamma_d.
\end{equation}
In particular, $m_0=\gamma_d$ and $m_1=\mu_N$. For $s,t\in[0,1]$, the
measure $N^{-1}\sum_i(F_s^i,F_t^i)_\#\gamma_d$ couples $m_s$ and $m_t$.
Its transport cost yields
\[
 W_2(m_s,m_t)^2
 \leq\frac1N\sum_i\int_{\R^d}|F_t^i(z)-F_s^i(z)|^2g(z)\,\dd z
 =|t-s|^2\left(d+\frac1N\sum_i|y_i|^2\right).
\]
Thus $m$ is Lipschitz in $W_2$ and belongs to
$\operatorname{AC}^2([0,1];\Pc_2(\R^d))$.

\medskip\noindent
\textbf{Step 2: The component densities and their fluxes.}
For $t<1$, the change of variables $x=F_t^i(z)$ gives
\[
 (F_t^i)_\#\gamma_d(\dd x)=K_t(x,y_i)\,\dd x,
 \qquad
 K_t(x,y_i)=(1-t)^{-d}
 g\!\left(\frac{x-ty_i}{1-t}\right).
\]
This is the Gaussian kernel in \eqref{eq:fm-marginal-velocity}, and
\eqref{eq:app-fm-mixture} gives the stated positive smooth density.
The velocity of the $i$th affine interpolation is
\[
 a_t^i(x)=\frac{y_i-x}{1-t},
 \qquad a_t^i(F_t^i(z))=y_i-z.
\]
For $\varphi\in C_c^\infty(\R^d)$, differentiation under the integral and
the same change of variables yield
\begin{align*}
 \frac{\dd}{\dd t}\int_{\R^d}\varphi(x)K_t(x,y_i)\,\dd x
 &=\int_{\R^d}\langle\nabla\varphi(F_t^i(z)),y_i-z\rangle
                  g(z)\,\dd z\\
 &=\int_{\R^d}\langle\nabla\varphi(x),a_t^i(x)\rangle
                  K_t(x,y_i)\,\dd x.
\end{align*}
The bound $\|\nabla\varphi\|_\infty(|y_i|+|z|)g(z)$ is integrable and
justifies the differentiation. Each component therefore satisfies a
continuity equation. Summing their fluxes gives
\[
 J_t(x)=\frac1N\sum_i a_t^i(x)K_t(x,y_i),
 \qquad p_t(x)=\frac{J_t(x)}{m_t(x)}
             =\frac{\bar y_t^F(x)-x}{1-t}.
\]
The positive Gaussian denominator implies continuity of $p$ on
$[0,1)\times\R^d$.

\medskip\noindent
\textbf{Step 3: The continuity equation.}
Summing the component equations and integrating in time shows that
every $\psi\in C_c^\infty((0,1)\times\R^d)$ satisfies
\[
 \int_0^1\!\int_{\R^d}
 \bigl(\partial_t\psi+\langle p_t,\nabla_x\psi\rangle\bigr)
 m_t(x)\,\dd x\,\dd t=0.
\]
The velocity also has finite kinetic action. Indeed, weighted
Cauchy--Schwarz gives
\[
 \int_{\R^d}|p_t(x)|^2m_t(x)\,\dd x
 \leq\frac1N\sum_i\int_{\R^d}|a_t^i(x)|^2K_t(x,y_i)\,\dd x
 =d+\frac1N\sum_i|y_i|^2.
\]
This proves the distributional equation with a velocity in
$L^2(m_t\,\dd t)$; see also \citet[Theorem~8.3.1]{AGS08}.
\end{proof}

\subsection{Exact trajectory correspondence}\label{sec:proof-equivalence}
\begin{proof}[Proof of Proposition~\ref{prop:exact-equivalence}]
\leavevmode\par\smallskip\noindent
\textbf{Step 1: Gaussian kernel identities.}
For $0<t<1$, direct substitution of $h(t)=(1-t)^2/(2t^2)$ gives
\[
 K_t(x,y_i)=t^{-d}G_{h(t)}(x/t-y_i).
\]
Summing these identities proves \eqref{eq:fm-heat-density}; taking their
weighted ratio gives $\bar y_t^F(x)=\bar y_{h(t)}^S(x/t)$. Since
$\bar y_h^S(z)=z+2h\,s(z,h)$,
\[
 p_t(x)
 =\frac{x/t+2h(t)s(x/t,h(t))-x}{1-t}
 =\frac{x}{t}+\frac{1-t}{t^2}s(x/t,h(t)).
\]
This is \eqref{eq:exact-field-correspondence}.

\medskip\noindent
\textbf{Step 2: From flow matching to heat probability flow.}
Fix $t_0\in(0,1)$ and let $T=h(t_0)$. For a flow-matching trajectory,
the preceding identity and $h'(t)=-(1-t)/t^3$ give
\[
 \frac{\dd}{\dd t}\left(\frac{X_t^F}{t}\right)
 =\frac{p_t(X_t^F)}{t}-\frac{X_t^F}{t^2}
 =-h'(t)s(X_t^F/t,h(t)).
\]
Consequently,
$X_r^S=X_{h^{-1}(r)}^F/h^{-1}(r)$ solves
$\dd X_r^S/\dd r=-s(X_r^S,r)$ on $(0,T]$, with
$X_T^S=X_{t_0}^F/t_0$. Spatial Lipschitz continuity and linear growth
on every positive-time strip give uniqueness throughout this interval.

\medskip\noindent
\textbf{Step 3: The converse.}
Let $X^S$ solve the heat probability flow with $X_T^S=x_T$, and set
$t_0=h^{-1}(T)$. For $t\in[t_0,1)$, define $X_t^F=tX_{h(t)}^S$.
Then
\begin{align*}
 \frac{\dd X_t^F}{\dd t}
 &=X_{h(t)}^S-th'(t)s(X_{h(t)}^S,h(t))\\
 &=\frac{X_t^F}{t}+\frac{1-t}{t^2}s(X_t^F/t,h(t))
 =p_t(X_t^F).
\end{align*}
At $t_0$ this trajectory equals $t_0x_T$. Uniqueness holds on every
$[t_0,b]$, $b<1$, and the same spatial bounds for $p$ on $[0,t_0]$
give a unique backward extension to zero. The inverse identity follows
from the bijectivity of $h$.
\end{proof}

\Needspace{5\baselineskip}
\subsection{The limiting theorem and its microscopic and macroscopic consequences}
\label{sec:proof-main}\label{sec:recalled-score}

We begin with the heat probability flow, since its small-noise limit determines the limiting flow-matching field through Proposition \ref{prop:exact-equivalence}. In logarithmic time, write
\begin{equation*}
 Y_\sigma^S=X_{Te^{-\sigma}}^S,\qquad
 \dot Y_\sigma^S=b_{Te^{-\sigma}}(Y_\sigma^S),\qquad
 b_h(x)=h\,s(x,h)=\tfrac12(\bar y_h^S(x)-x).
\end{equation*}

The single-score specialization of
\citet[Theorems~2.1 and~2.5]{liuzuazua2026convergence} gives complete graphical
convergence $b_h\to-\partialC\Phi_N/4$ as $h\to0^+$, locally uniform
convergence off $\Sigma_N$, and the same limiting inclusion for time-shift
limits. Every deterministic trajectory converges to an equilibrium. By
\citet[Theorem~3.7(i)]{liuzuazua2025pde}, Lebesgue-almost every initial state has an endpoint in $\Y$. Such endpoints lie in strict cells, so the estimate of \citet[Theorem~2.5(ii)]{liuzuazua2026convergence} gives $|X_h^S-y^{S,*}|=O(\sqrt h)$. If the initial law is $u_T$, the exact marginals $u_h$ converge weakly to $\mu_N$, which therefore is the endpoint law. 

\begin{proof}[Proof of Theorem~\ref{thm:main}]
\leavevmode\par\smallskip\noindent
\textbf{Step 1: The limiting field.}
Proposition~\ref{prop:exact-equivalence} gives
\begin{equation*}
 B_t(x)=b_{h(t)}(x/t)+\frac{1-t}{2t}x.
\end{equation*}

The correction vanishes locally uniformly, while $h(t)\to0^+$ and $x/t\to x$.
More explicitly, if $t_n\to1^-$, $x_n\to x$, and $B_{t_n}(x_n)\to v$, then
\[
 b_{h(t_n)}(x_n/t_n)
 =B_{t_n}(x_n)-\frac{1-t_n}{2t_n}x_n\longrightarrow v.
\]

The upper heat graphical limit gives $v\in-\partialC\Phi_N(x)/4$.
Local uniform convergence off $\Sigma_N$ follows by keeping $x/t$ in a
compact neighborhood with unique projections. Conversely, for
$v\in-\partialC\Phi_N(x)/4$ and any $t_n\to1^-$, choose heat recovery states
$z_n\to x$ with $b_{h(t_n)}(z_n)\to v$. Then $x_n=t_nz_n\to x$ and
\[
 B_{t_n}(x_n)=b_{h(t_n)}(z_n)+\frac{1-t_n}{2}z_n\longrightarrow v.
\]
This proves complete graphical convergence.

\medskip\noindent
\textbf{Step 2: Time-shift limits.}
For each fixed initial state, the matched heat trajectory has an
equilibrium endpoint $y^*$. Proposition~\ref{prop:exact-equivalence} gives,
for all sufficiently large $\tau$,
\[
 Y_\tau^F=t_F(\tau)X_{h(t_F(\tau))}^S\longrightarrow y^*.
\]
For every $S>0$ and every sequence $\tau_n\to\infty$, it follows that
\[
 \sup_{0\leq s\leq S}|Y_{\tau_n+s}^F-y^*|
 \leq\sup_{\sigma\geq\tau_n}|Y_\sigma^F-y^*|
 \longrightarrow0.
\]
Thus the time translates converge locally uniformly to the constant path
$y^*$, and $0\in\partialC\Phi_N(y^*)$ makes this path a solution of
\eqref{eq:limiting-inclusion}. This proves (i). The prescribed restarts
in part (ii) instead keep the initial state fixed while moving the
starting time toward the terminal regime.

\medskip\noindent
\textbf{Step 3: Uniform convergence for all elapsed times.}
Let \(\mathcal V_i\) denote the open Voronoi cell of \(y_i\), and consider the set of full reference segments
\[
\mathcal C_K = \bigcup_{i=1}^N \Bigl\{y_i+a(x-y_i):x\in K\cap\mathcal V_i,\ a\in[0,1]\Bigr\}.
\]

Since \(K\) is compact and avoids \(\Sigma_N\), each \(K\cap\mathcal V_i\) is closed in \(K\) and hence compact. By convexity of \(\mathcal V_i\), each segment from \(x\in K\cap\mathcal V_i\) to \(y_i\) lies in \(\mathcal V_i\). Thus \(\mathcal C_K\) is compact, is disjoint from \(\Sigma_N\), and contains \(F_s(x)\) for all \(x\in K\) and \(s\geq0\). Choose \(r>0\) so that the closed \(r\)-neighborhood \(A\) of \(\mathcal C_K\) is disjoint from \(\Sigma_N\). 

If \(N=1\), the exact and reference fields coincide, so \eqref{eq:finite-window-limit} holds. Assume \(N\ge 2\), and let \(i(z)\) be the unique nearest-site index of \(z\in A\). Since \(A\) is compact and disjoint from the tie set, the gap between the smallest and second-smallest squared distances has a positive minimum,
\[
\delta_A = \min_{z\in A}\min_{j\ne i(z)} \Bigl(|z-y_j|^2-|z-y_{i(z)}|^2\Bigr)>0.
\]

By compactness of \(A\) and finiteness of \(\Y\), there exists \(t_A<1\) such that, for every \(t\in[t_A,1)\), \(z\in A\), and \(j\ne i(z)\), \(|z-t y_j|^2-|z-t y_{i(z)}|^2\ge \tfrac{\delta_A}{2}\). Let \(w_j(t,z)\) denote the normalized Gaussian weight of \(y_j\) in \(\bar y_t^F(z)\). Bounding the denominator below by its \(i(z)\)-th term gives

\[
w_j(t,z)\le
\exp\!\left(-\frac{\delta_A}{4(1-t)^2}\right)
\qquad\text{for }j\ne i(z).
\]
Since the Gaussian weights sum to one, the difference between the exact and reference fields is
\[
B_t(z)-b_N(z)
=\frac12\sum_{j\ne i(z)}w_j(t,z)\bigl(y_j-y_{i(z)}\bigr).
\]
Consequently, using \(1-t_F(\tau)=e^{-\tau/2}\), for every \(\ell\) such that \(t_F(\ell)\ge t_A\),
\[
\varepsilon_\ell :=\sup_{\tau\ge\ell}\sup_{z\in A} 
\left|B_{t_F(\tau)}(z)-b_N(z)\right|
\le
\frac{(N-1)\operatorname{diam}(\Y)}{2}
\exp\!\left(-\frac{\delta_A}{4}e^\ell\right).
\]

Fix \(x\in K\) and set \(E_s=Z_s^{\ell,x}-F_s(x)\). As long as \(|E_s|<r\), the segment joining \(F_s(x)\) to \(Z_s^{\ell,x}\) lies in \(A\), so both points belong to the same Voronoi cell. The affine form of \(b_N\) on that cell gives
\[
\dot E_s=-\tfrac12 E_s+R_s,\qquad E_0=0,
\qquad
R_s=B_{t_F(\ell+s)}(Z_s^{\ell,x})-b_N(Z_s^{\ell,x}),
\]
with \(|R_s|\leq\varepsilon_\ell\). Variation of constants yields
\[
|E_s|
\leq \varepsilon_\ell\int_0^s e^{-(s-u)/2}\,du
=2\varepsilon_\ell(1-e^{-s/2})
\leq2\varepsilon_\ell.
\]

The bound is independent of the elapsed time because the affine
contraction has an integrable kernel. All constants were chosen uniformly
on the compact tube $A$.

For sufficiently large \(\ell\), we have \(2\varepsilon_\ell<r\). If \(|E_s|\) first reached \(r\) at a finite time \(s_*\), continuity would give \(r=|E_{s_*}|\leq2\varepsilon_\ell<r\), a contradiction. The estimate therefore holds for every \(s\geq0\), uniformly over \(x\in K\). In particular,
\[
\sup_{x\in K}\sup_{s\ge0}
\left|Z_s^{\ell,x}-F_s(x)\right|
\le 2\varepsilon_\ell
\le (N-1)\operatorname{diam}(\Y)
\exp\!\left(-\frac{\delta_A}{4}e^\ell\right)\longrightarrow 0,
\qquad\text{as }\ell\to\infty.
\]
This proves \eqref{eq:finite-window-limit}.
\end{proof}

\begin{proof}[Proof of Corollary~\ref{cor:fm-endpoints}]
Use Proposition~\ref{prop:exact-equivalence} with $t_0=1/2$.
The identity $X_t^F=tX_{h(t)}^S$ transfers every deterministic heat endpoint
to the same flow-matching endpoint. Under Gaussian initialization, the
matched heat initial law is $u_{1/2}$; hence its endpoint is a training site
$y_i$ almost surely and has law $\mu_N$. The heat estimate gives
\[
 |X_t^F-y_i|\leq t|X_{h(t)}^S-y_i|+(1-t)|y_i|
 \leq(C/\sqrt2+|y_i|)(1-t).
\]
Substitution of $1-t_F(\tau)=e^{-\tau/2}$ gives the second rate.
See also \citet[Theorems~5.3--5.4]{wan2025denoisers}.
\end{proof}

\begin{proof}[Proof of Corollary~\ref{cor:fm-macroscopic}]
\leavevmode\par\smallskip\noindent
\textbf{Step 1: A global coupling bound.}
Write $D_\Y=\operatorname{diam}(\Y)$ and define
\[
 \mathcal E_q(\ell)^2
 :=\int_{\R^d}\sup_{s\geq0}
       |\mathcal Z_s^\ell(x)-F_s(x)|^2\,q(\dd x).
\]

Both flows have linear part $-Z/2$ and forcing in $\tfrac12\conv\Y$.
Variation of constants gives, for every $x\notin\Sigma_N$,
\[
 |\mathcal Z_s^\ell(x)-F_s(x)|
 \leq\frac{D_\Y}{2}\int_0^s e^{-(s-u)/2}\,\dd u
 =D_\Y(1-e^{-s/2}).
\]

Theorem~\ref{thm:main}(ii) with $K=\{x\}$ and dominated convergence
give $\mathcal E_q(\ell)\to0$. Both laws have finite second
moments, since each trajectory is bounded by
$e^{-s/2}|x|+(1-e^{-s/2})\max_i|y_i|$.
Coupling them through the same initial point yields
\[
 \sup_{s\geq0}W_2\big((\mathcal Z_s^\ell)_\#q,(F_s)_\#q\big)
 \leq\mathcal E_q(\ell),
\]
which proves the qualitative convergence.
For $N=1$ the two flows coincide. We henceforth assume $N\geq2$.

\medskip\noindent
\textbf{Step 2: Uniform control outside a shrinking interface layer.}
Write $m=\min_{i\ne j}|y_i-y_j|>0$, $R=\max_i|y_i|$,
$a=e^{-\ell/2}$, $C_0=4R+1$, and $r=C_0a$.
Set $\Sigma_N^r=\{x:\dist(x,\Sigma_N)\leq r\}$.
Take $\ell$ large enough that $a\leq1/2$ and $r\leq m/2$.
For $x\in\mathcal V_i$ with $\dist(x,\Sigma_N)>r$, introduce the
signed distance to each cell-defining bisector,
\[
 d_{ij}(z)=\frac{|z-y_j|^2-|z-y_i|^2}{2|y_j-y_i|},\qquad j\ne i.
\]

The distance from $x$ to the complement of $\mathcal V_i$ is its distance to the boundary. The complement is the union of the closed half-spaces $\{d_{ij}\le0\}$, so $d_{ij}(x)>r$ for every $j\ne i$, while $d_{ij}(y_i)=|y_j-y_i|/2\geq r$.
By affinity, $d_{ij}(F_s(x))\geq r$ for all $s\geq0$.
Thus $|z-F_s(x)|\leq r/2$ implies $d_{ij}(z)\geq r/2$.
For $t=t_F(\ell+s)$, the scaled-center gap satisfies
\begin{align*}
 |z-ty_j|^2-|z-ty_i|^2
 &=2t|y_j-y_i|\left(d_{ij}(z)
       -(1-t)\frac{|y_j|^2-|y_i|^2}{2|y_j-y_i|}\right)\\
 &\geq2t|y_j-y_i|(r/2-Ra)\geq mr/4.
\end{align*}
The Gaussian-weight estimate from the proof of
Theorem~\ref{thm:main} therefore gives
\[
 |B_t(z)-b_N(z)|
 \leq\frac{(N-1)D_\Y}{2}
           \exp\left(-\frac{mr}{8a^2}\right).
\]
The same variation-of-constants and first-exit argument yields
\[
 \sup_{s\geq0}|\mathcal Z_s^\ell(x)-F_s(x)|
 \leq(N-1)D_\Y\exp\left(-\frac{mC_0}{8a}\right)
 \qquad\bigl(x\notin\Sigma_N^r\bigr).
\]

Indeed, for all sufficiently large $\ell$ the right-hand side is smaller
than $r/2$, so the trajectory stays in the same radius-$r/2$ tube for
all elapsed times. These constants are uniform in $x$, including
unbounded initial states. Splitting the integral over $\Sigma_N^r$ and
its complement, and using the global diameter bound, gives
\[
 \mathcal E_q(\ell)^2
 \leq D_\Y^2\,q\big(\Sigma_N^{C_0e^{-\ell/2}}\big)
       +C_1\exp\big(-c_1e^{\ell/2}\big),
\]
where $C_1=(N-1)^2D_\Y^2$ and $c_1=mC_0/4$.
In particular, if $q(\Sigma_N^r)\leq A_qr^\alpha$ for all sufficiently
small $r>0$, with $A_q,\alpha>0$, then
$\mathcal E_q(\ell)=O(e^{-\alpha\ell/4})$, and the same rate holds
for the all-time Wasserstein error.

\medskip\noindent
\textbf{Step 3: Gaussian sources and sharpness.}
The tie set lies in the union of the $N(N-1)/2$ perpendicular bisectors.
If the one-dimensional marginals normal to these hyperplanes have
densities bounded by $M_q$, a union bound gives
\[
 q(\Sigma_N^r)\leq N(N-1)M_q r.
\]

For $\gamma_d$, take $M_q=(2\pi)^{-1/2}$; for $u_T$, take
$M_q=(4\pi T)^{-1/2}$; and for the fixed tube law, take
$M_q=(4\pi t_{\min})^{-1/2}$. This proves the stated rates.

To see sharpness, let $d=1$, $\Y=\{1,3\}$, and $q=\gamma_1$.
At $x=2t$ the two Gaussian weights are equal, so
\[
 \bar y_t^F(2t)=2,\qquad p_t(2t)=\frac{2-2t}{1-t}=2.
\]
Hence $x=2t$ is an exact trajectory. Uniqueness preserves order on every
time interval ending before $t=1$: a restart below this curve has endpoint
at most $2$, while a restart above it has endpoint at least $2$.
For an absolutely continuous restart at $t_\ell=t_F(\ell)>0$, the matched
heat restart also has a density. Almost every endpoint is therefore a
site, giving $1$ below $2t_\ell$ and $3$ above it. The reference contraction
instead separates its basins at $2$.

Let $\Psi$ and $\psi$ be the standard Gaussian distribution function and
density. The exact and reference endpoint laws are, respectively,
\[
 \eta_\ell=\Psi(2t_\ell)\delta_1+
              [1-\Psi(2t_\ell)]\delta_3,
 \qquad
 \eta_*=\Psi(2)\delta_1+[1-\Psi(2)]\delta_3.
\]
Their quadratic transport cost is
$4[\Psi(2)-\Psi(2t_\ell)]$, since precisely this excess mass must cross
the distance between the two sites. The global trajectory bound in
Step~1 gives convergence to the endpoint laws in $W_2$ by dominated
convergence. Consequently,
\[
 \sup_{s\geq0}W_2^2\big((\mathcal Z_s^\ell)_\#\gamma_1,
                       (F_s)_\#\gamma_1\big)
 \geq4\bigl[\Psi(2)-\Psi(2t_\ell)\bigr]
 \sim8\psi(2)e^{-\ell/2}.
\]
Together with the upper bound, this establishes sharpness of the
Wasserstein exponent $1/4$.
\end{proof}

\subsection{Path-weighted transport estimate}\label{sec:proof-transport}
\begin{proof}[Proof of Proposition~\ref{prop:weighted-transport}]
\leavevmode\par\smallskip\noindent
\textbf{Step 1: One-sided stability.}
The regularity assumptions give local existence and uniqueness. Set
$M_0=\sup_{0\leq\tau\leq\tau_f}|V_\theta(0,\tau)|$.
By \eqref{eq:one-sided-bound} and Young's inequality,
\[
 \frac12\frac{\dd}{\dd\tau}|Z_\tau|^2
 \leq\left(\lambda+\frac12\right)|Z_\tau|^2+\frac12M_0^2.
\]

Gronwall's inequality gives global existence on $[0,\tau_f]$ and
preservation of finite second moments.
Use the same source $\widehat Z_0=X_0\sim q_{\mathrm{tube}}$ for
the learned trajectory and $A_s=F_s(X_0)$. Put
$E_s=\widehat Z_s-A_s$ and
$R_s=V_\theta(A_s,s)-b_N(A_s)$. Then, almost everywhere,
\[
 \frac12\frac{\dd}{\dd s}|E_s|^2
 \leq\lambda|E_s|^2+|E_s||R_s|.
\]
Set $e_\eps(s)=(|E_s|^2+\eps^2)^{1/2}$. The preceding
inequality implies
\[
 \dot e_\eps(s)
 \leq\lambda e_\eps(s)+|R_s|+|\lambda|\eps,
 \qquad e_\eps(0)=\eps.
\]
Multiplication by $e^{-\lambda s}$, integration, and passage to
$\eps\downarrow0$ yield
\[
 |E_{\tau_f}|\leq\int_0^{\tau_f}
                 e^{\lambda(\tau_f-s)}|R_s|\,\dd s.
\]

The pair $(\widehat Z_{\tau_f},A_{\tau_f})$ couples the learned and
reference laws. Since
$\int_0^{\tau_f}\E|R_s|^2\,\dd s=\mathcal L_{\mathrm{PDFM}}(\theta)$,
Minkowski's and Cauchy-Schwarz inequalities give
\begin{equation*}
 W_2(\Law(\widehat Z_{\tau_f}),(F_{\tau_f})_\#q_{\mathrm{tube}}) 
 \leq\int_0^{\tau_f} e^{\lambda(\tau_f-s)}
                     (\E|R_s|^2)^{1/2}\,\dd s
 \leq\sqrt{\kappa_\lambda(\tau_f)
                 \mathcal L_{\mathrm{PDFM}}(\theta)}. 
\end{equation*}

\medskip\noindent
\textbf{Step 2: Exact contraction and allocation.}
For almost every $X_0\sim q_{\mathrm{tube}}$, cell invariance gives
\[
 \dist(F_\tau(X_0),\Y)=|F_\tau(X_0)-P_\Y(X_0)|
                      =e^{-\tau/2}\dist(X_0,\Y).
\]

Coupling $F_\tau(X_0)$ with $P_\Y(X_0)$ gives
\[
 W_2^2\big((F_\tau)_\#q_{\mathrm{tube}},\nu_*\big)
 \leq e^{-\tau}D_{\mathrm{tube}}^2.
\]
Conversely, every coupling to a measure supported on $\Y$ has cost at
least the integral of squared distance to $\Y$. For the first marginal
$(F_\tau)_\#q_{\mathrm{tube}}$, that integral equals
$e^{-\tau}D_{\mathrm{tube}}^2$. This proves the equality
\eqref{eq:hard-flow-wasserstein-identity}. The triangle inequality through the
finite-time contraction law and $\nu_*$ supplies the remaining contraction
and allocation terms. Since $\Law(\widehat Z_{\tau_f})=\nu_{\theta,\tau_f}$, combining the two
steps proves \eqref{eq:error-separation}.
\end{proof}

\section{Endpoint allocation and numerical experiments}\label{sec:numerical-experiments}
We first illustrate the role of a moving interface in a two-site example.
The MNIST study then compares PDL and PDFM with denoising score matching
(DSM), heat-noise prediction, and linear flow matching. The five methods
use the same training support, update budget, and generation states.
The evaluation follows the geometry of the analysis: proximity to the
support, allocation among its sites, and fidelity to held-out images.

\subsection{Allocation error at a two-site interface}
For $\Y=\{1,3\}$ and a standard Gaussian restart, the exact flow has
basin boundary $2t_\ell$, whereas the reference contraction has boundary
$2$, with $t_\ell=1-e^{-\ell/2}$. Let $\Psi$ denote the standard Gaussian
distribution function. The interval between these boundaries carries mass
$\delta_\ell=\Psi(2)-\Psi(2t_\ell)$. Its reassignment between sites at
distance two gives the exact endpoint discrepancy
\[
 W_2(\eta_\ell,\eta_*)=2\sqrt{\delta_\ell},\qquad
 e^{\ell/4}W_2(\eta_\ell,\eta_*)\longrightarrow
 \sqrt{8\Psi'(2)}.
\]
Here $\eta_\ell$ and $\eta_*$ are the exact and reference endpoint laws
from the proof of Corollary~\ref{cor:fm-macroscopic}. This calculation
identifies how a shrinking layer of initial states produces the
macroscopic exponent. The reference path in either strict cell is
$F_s(x)=y_i+e^{-s/2}(x-y_i)$. On compact subsets of the cell,
Theorem~\ref{thm:main}(ii) controls the entire exact trajectory by the
much smaller cellwise error.

\subsection{Data, training, and generation}\label{sec:datasets-targets}
MNIST images are represented in $[0,1]^{784}$~\citep{lecun1998gradient}.
The support contains $N=10{,}000$ training images, with $1{,}000$ per digit,
selected with seed $123$. All nearest-site targets search this full support.
Tube queries use $Q=D+\sqrt{2t}\,G$, where $D\sim\mu_N$, $G\sim\gamma_d$,
and $\log t$ is uniform on $[\log t_{\min},\log T]$, with
$T=1/2$ and $t_{\min}=Te^{-12}$. The fixed query pool contains
$120{,}000$ states, twelve per support image.

All models are trained for $100{,}000$ updates with batch size $64$.
AdamW uses peak learning rate $2\times10^{-4}$, weight decay $10^{-5}$,
gradient clipping at one, a $500$-update warmup, and cosine decay to one
tenth of the peak rate. Generation uses the final exponential-moving-average
weights, with decay $0.999$. The residual U-Nets have base width $64$,
three spatial resolutions, and attention at $7\times7$. The autonomous
scalar PDL network contains $4{,}295{,}809$ parameters; each time-conditioned
vector network contains $2{,}993{,}281$. Their layer specifications and
conditioning conventions are given in Appendix~\ref{supp:implementation}.

\paragraph{Training targets.}
Writing $e_\theta=\Phi_\theta-\Phi_N$ and $d=784$, the implemented PDL
loss is
\[
 \E\left[\left|\frac{e_\theta(Q)}d\right|^2
             +\frac1d|\nabla e_\theta(Q)|^2\right].
\]
It is $d^{-2}\mathcal L_{\mathrm{PDL}}^\beta$ with $\beta=d$ in
\eqref{eq:pdl-augmented-loss}. PDFM samples $\tau$ uniformly on $[0,12]$
along \eqref{eq:exp-pdfm-path} and fits $V_\tau^*$ by mean squared error
per pixel. DSM fits the rescaled heat drift $t\,s_\theta(Q,t)$ to
$(D-Q)/2$, while heat-noise prediction fits $\epsilon_\theta(Q,t)$ to
$G$~\citep{vincent2011connection,ho2020ddpm}. Their generation velocities
are, respectively, the fitted drift and
$-\sqrt{t/2}\,\epsilon_\theta$. Linear FM fits $v_\theta(C_r,r)$ to
$D-G$, using independent $D,G$ and uniform $r\in[0,1]$
in \eqref{eq:straight-fm-path}~\citep{lipman2023flow}. All vector losses
are averaged over pixels.

\paragraph{Common noisy initialization.}
Each method evolves the same saved $M=10{,}000$ initial states
\begin{equation}\label{eq:numerical-source}
 X_0=D+G\sim u_{1/2},\qquad D\sim\mu_N,\quad G\sim\gamma_d,
\end{equation}
drawn with seed $1123$. Generation uses logarithmic heat time
$t(\tau)=Te^{-\tau}$ on $0\leq\tau\leq12$ and $480$ explicit Euler
steps of size $0.025$. Thus the terminal heat time equals the smallest
training noise time. PDL follows its potential gradient, and PDFM
evaluates its learned path velocity.

To use these same physical states for FM, apply the space-time change
from Section~\ref{sec:exact-equivalence}. For the native learned trajectory
$\dd Z_r/\dd r=v_\theta(Z_r,r)$, define
\begin{equation}\label{eq:numerical-fm-heat}
 r(\tau)=\frac{1}{1+\sqrt{2T}e^{-\tau/2}},\qquad
 X_\tau=\frac{Z_{r(\tau)}}{r(\tau)},\qquad
 \dot X_\tau=\frac{1-r(\tau)}2
       \big[v_\theta(r(\tau)X_\tau,r(\tau))-X_\tau\big].
\end{equation}
The saved FM field is thereby evaluated from native time $r(0)=1/2$
to $r(12)\simeq0.997527$, with the common initialization
\eqref{eq:numerical-source} and terminal heat time. The logarithmic heat
clock used here differs from $t_F(\tau)$ in the late-restart theorem;
\eqref{eq:numerical-fm-heat} specifies their sampling correspondence.
Numerical trajectories are stored as raw floating-point states.
Displayed images and image-quality metrics use endpoint clipping to
$[0,1]$; geometric diagnostics use the raw states.

\subsection{Proximity, occupancy, and feature fidelity}\label{sec:evaluation-protocol}
For each clipped output, a pixelwise Euclidean search identifies its
nearest support image. NN-MSE is the mean squared difference per pixel;
NN-SSIM~\citep{wang2004ssim} and NN-LPIPS~\citep{zhang2018lpips} compare
the same pair. SSIM uses an $11\times11$ Gaussian window with standard
deviation $1.5$. LPIPS uses AlexNet after replicating grayscale over three
channels and resizing to $64\times64$. These metrics describe local
agreement with the empirical support.

Let $n_i$ be the number of outputs assigned to support image $y_i$, and
put $\widehat p_i=n_i/M$. We report
\[
 \mathrm{Coverage}=\frac{\#\{i:n_i>0\}}N,\qquad
 \mathrm{Entropy}=-\frac{\sum_i\widehat p_i\log\widehat p_i}{\log N},
 \qquad 0\log0=0.
\]
Coverage records the fraction of visited sites, while entropy records
the spread of their empirical masses. Even uniform independent sampling
of $N$ sites visits only an expected fraction
$1-(1-1/N)^M$, approximately $0.6321$ when $M=N=10{,}000$.
The labeled masses, together with distances between sites, determine
the allocation term in Proposition~\ref{prop:weighted-transport}.

Feature FID~\citep{heusel2017fid} and KID~\citep{binkowski2018kid}
are computed separately against the $10{,}000$ support images and all
$10{,}000$ MNIST test images. Every method uses the same frozen
$128$-dimensional MNIST classifier embedding, trained on the standard
$60{,}000$-image training split. These are MNIST-feature scores.
KID uses the kernel $(1+\langle z,z'\rangle/128)^3$ and averages
$100$ unbiased estimates on subsets of $1{,}000$ images per distribution,
with shared subset indices and seed $2123$. It is reported on its raw
scale. Appendix~\ref{supp:implementation} specifies the encoder and numerical
metric conventions.

Table~\ref{tab:main-results} brings together the local, occupancy, and
feature measurements. The training and test references distinguish
agreement with the fitted support from agreement with held-out images.
FM attains the smallest NN-MSE and NN-LPIPS and the largest NN-SSIM.
PDL ranks second on these three measures: its NN-MSE is $9.1\%$ below
DSM and $11.6\%$ below heat-noise prediction. PDL also has the highest
measured coverage and occupancy entropy, visiting $6{,}256$ support
images, compared with $6{,}222$ for DSM and $6{,}200$ for FM.

The feature scores give a different ordering. FM has the lowest training
and test FID and KID, followed by DSM. PDL has the largest values in
these four columns; its test FID is $18.891$, compared with $3.734$ for
FM and $7.685$ for DSM. Thus proximity to individual training images
and agreement with the distribution of image features capture distinct
properties of the learned dynamics.

\begin{figure}[!p]
\centering
\includegraphics[width=.8\linewidth]{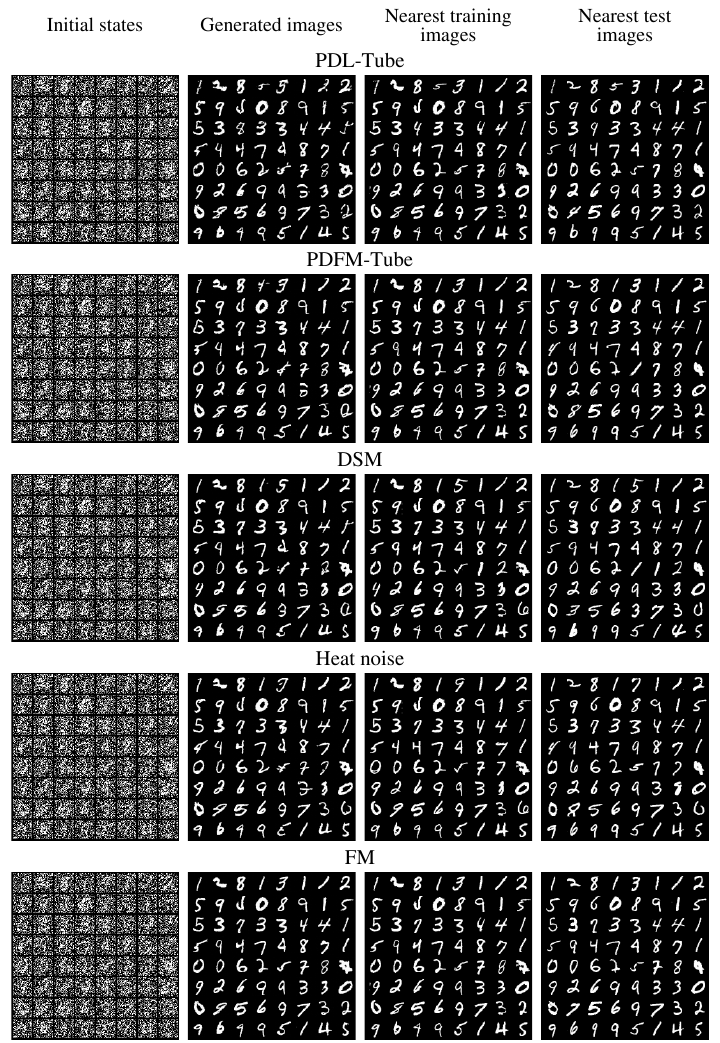}
\caption{MNIST generation from common noisy-data initial states.
Rows correspond to the five methods. Columns show initial states,
outputs at $\tau=12$, and pixel-nearest training and test images.
Each panel contains the first $64$ samples in saved order; a fixed
grid position identifies the same initial state throughout the figure.
Images are clipped to $[0,1]$ for display.}
\label{fig:comparison}
\end{figure}

Figure~\ref{fig:comparison} compares the first $64$ samples in their
common saved order with their nearest training and test images.
The outputs recover recognizable digits from the noisy initial states,
with differences in stroke shape and continuity. Appendix~\ref{supp:images} displays the corresponding generation trajectories.

\input{tables/main_results.tex}
\FloatBarrier

\subsection{Reference geometry and endpoint assignments}\label{sec:geometric-diagnostics}
For raw generated states $Z_j$, let $X_{0,j}$ denote the common starts.
The quantities
\[
 E_{\Y}=\frac1{Md}\sum_{j=1}^M\dist(Z_j,\Y)^2,
 \qquad
 A=\frac1M\sum_{j=1}^M
 \mathbf1_{\{P_\Y(Z_j)=P_\Y(X_{0,j})\}}
\]
measure support proximity and retention of the initial nearest-site
assignment. Exact contractions $F_{12}(X_{0,j})$ and nearest-site
projections $P_\Y(X_{0,j})$ provide geometric references under the same
source law. Cell invariance gives $A=1$ for both references; for the
finite-time contraction,
\[
 \E\frac{\dist(F_{12}(X_0),\Y)^2}{d}
 =e^{-12}\E\frac{\dist(X_0,\Y)^2}{d}
 \leq 2Te^{-12}.
\]
The last inequality uses $X_0=D+\sqrt{2T}G$ and
$\dist(X_0,\Y)\leq|X_0-D|$.
Table~\ref{tab:geometry} records these quantities. PDL retains $76.03\%$
of the initial nearest-site assignments, compared with $73.33\%$ for FM,
$71.91\%$ for PDFM, $63.82\%$ for DSM, and $62.34\%$ for heat-noise
prediction. FM has the smallest raw support MSE, followed by PDL.
The empirical reference contraction gives $E_{\Y}=6.136\times10^{-6}$,
in agreement with the exact contraction identity. The learned endpoints
remain farther from the support than this geometric reference.

\input{tables/geometry_results.tex}

The projection and contraction controls have test feature FID $4.598$
and $4.602$, respectively; their complete feature scores are included in
the accompanying source archive. These controls connect the explicit reference
maps with image-distribution measurements. Taken together, the two tables
show that PDL preserves the initial nearest-site geometry more often,
whereas FM produces closer support matches and stronger feature fidelity.
The reference cell masses, the accuracy of the learned field, and the
resulting image features each contribute to the observed output law.
\FloatBarrier

\section{Conclusions}\label{sec:conclusion}

The geometry of a finite training set gives an explicit description of
terminal flow-matching dynamics. Within a strict Voronoi cell, the
reference motion contracts exponentially toward its training site.
An exact trajectory restarted at a prescribed late-time state follows
this contraction throughout its remaining evolution, with an error
exponentially small in the inverse squared noise scale. At ties,
graphical convergence identifies the full Clarke subgradient field.
The correspondence with heat probability flow places both generative
equations in the same geometric setting.

For distributions of starting states, a thin layer near a moving
interface can change its endpoint assignment. This mechanism gives the
square-root dependence of the Wasserstein error on the restart noise
scale for Gaussian and fixed Gaussian-tube laws. The two-site example
attains this exponent and makes the effect of reallocation explicit.

PDL and PDFM turn this geometry into supervision for a distance
potential and for velocities along explicit paths. Under a one-sided
expansion condition, the transport estimate relates velocity accuracy to
the output law, together with finite generation time and endpoint
allocation. Since the initial law determines the reference cell masses,
source design and velocity learning are complementary parts of the
construction.

The MNIST study compares five learned fields from common noisy initial
states. PDL gives the highest measured nearest-site retention and support
occupancy, and ranks second in proximity to the training support. FM
achieves the strongest image and feature-distribution fidelity. These
findings distinguish the preservation of nearest-site geometry from
agreement with an image distribution. They also motivate learning
initializations with prescribed cell masses and evaluating velocity
accuracy along the paths followed during generation. Extending the
explicit reference-path approach to data supported on continuous
geometric sets is a further direction.

\Needspace{5\baselineskip}
\paragraph{Computational materials.}
Appendices~\ref{supp:implementation} and~\ref{supp:images} provide
network details and generation trajectories. The arXiv ancillary archive
contains code, configurations, numerical tables, figure assets, and
provenance information for the experiments.

\clearpage
\appendix
\numberwithin{figure}{section}
\numberwithin{table}{section}
\input{appendices.tex}

\section*{Acknowledgments}
This project received funding from the European Research Council (ERC) under the European Union's Horizon Europe research and innovation programme (Grant Agreement No. 101096251-CoDeFeL). This material is based upon work supported by the Air Force Office of Scientific Research under Award No. FA8655-22-1-7012. EZ was partially supported by the Alexander von Humboldt Professorship program; the European Union's Horizon Europe MSCA project ModConFlex (HORIZON-MSCA-2021-DN-01, Project No. 101073558); the DFG Transregio 154 project, Mathematical Modeling, Simulation and Optimization Using the Example of Gas Networks; SURE-AI: The Norwegian Centre for Sustainable, Risk-Averse, and Ethical AI (Grant 357482, Research Council of Norway); and the Madrid Government-UAM Agreement for the Excellence of the University Research Staff in the context of the V PRICIT. UB and EZ were partially supported by the Basque Government through the consolidated research group ``CMRG-Computational Mathematics Research Group'' under Grant IT1856-26, by Grant PID2023-146872OB-I00-DyCMaMod of MICIU (Spain) and by COST Action CA24136-Interactions between Control Theory and Machine Learning. UB was partially supported by COST Action CA24122-Multiscale Stochastics, Patterns, and Analysis of Combinatorial Environments.

\paragraph{AI-assisted tools.}
AI-assisted tools were used in manuscript revision and consistency checking.
The authors assume responsibility for all content.

\FloatBarrier
\clearpage
\bibliographystyle{plainnat}
\bibliography{reference}
\end{document}

%% file: tables/main_results.tex
\begin{table}[!htbp]
\centering\small
\setlength{\tabcolsep}{5pt}
\textbf{(a) Nearest-support proximity and occupancy}\par\smallskip
\begin{tabular}{@{}lrrrrr@{}}
\toprule
Method & NN-MSE $\downarrow$ & NN-SSIM $\uparrow$ & NN-LPIPS $\downarrow$ & Coverage $\uparrow$ & Entropy $\uparrow$ \\
\midrule
PDL-Tube & \underline{0.02280} & \underline{0.8204} & \underline{0.07353} & \textbf{0.6256} & \textbf{0.93621} \\
PDFM-Tube & $0.02520$ & $0.8051$ & $0.07662$ & $0.6220$ & $0.93498$ \\
DSM & $0.02508$ & $0.8071$ & $0.07495$ & \underline{0.6222} & \underline{0.93508} \\
Heat noise & $0.02578$ & $0.8031$ & $0.07749$ & $0.6119$ & $0.93285$ \\
FM & \textbf{0.01633} & \textbf{0.8649} & \textbf{0.04970} & $0.6200$ & $0.93462$ \\
\bottomrule
\end{tabular}\par
\medskip
\textbf{(b) MNIST-feature fidelity}\par\smallskip
\begin{tabular}{@{}lrrrr@{}}
\toprule
Method & Train FID $\downarrow$ & Test FID $\downarrow$ & Train KID $\downarrow$ & Test KID $\downarrow$ \\
\midrule
PDL-Tube & $14.090$ & $18.891$ & $17.969$ & $18.004$ \\
PDFM-Tube & $6.525$ & $8.958$ & $9.044$ & $8.119$ \\
DSM & \underline{5.204} & \underline{7.685} & \underline{6.524} & \underline{5.554} \\
Heat noise & $5.678$ & $7.994$ & $7.275$ & $6.235$ \\
FM & \textbf{1.144} & \textbf{3.734} & \textbf{0.238} & \textbf{0.091} \\
\bottomrule
\end{tabular}\par
\caption{Results for $10{,}000$ generated images per method from the
same noisy empirical source. Training references are the $10{,}000$
support images; test references are all $10{,}000$ MNIST test images.
Feature scores use one frozen MNIST encoder; KID is on its raw scale.
All methods use their final EMA weights after $100{,}000$ updates and
$480$ Euler steps to $\tau=12$. FM is sampled in the common heat
coordinates. \textbf{Boldface} and \underline{underlining} indicate,
respectively, the best and second-best distinct values in each column,
ranked before rounding.}
\label{tab:main-results}
\end{table}

%% file: tables/geometry_results.tex
\begin{table}[!htbp]
\centering\small
\setlength{\tabcolsep}{16pt}
\begin{tabular}{@{}lrr@{}}
\toprule
Method & Raw support MSE $E_{\Y}\downarrow$ & Site retention $A\uparrow$\\
\midrule
PDL-Tube & \underline{$0.022823$} & \textbf{0.7603} \\
PDFM-Tube & $0.025270$ & $0.7191$ \\
DSM & $0.025125$ & $0.6382$ \\
Heat noise & $0.025799$ & $0.6234$ \\
FM & \textbf{0.016340} & \underline{$0.7333$} \\
\midrule
Exact contraction $F_{12}$ & $6.136\times10^{-6}$ & $1$ \\
Nearest-support projection $P_{\Y}$ & $0$ & $1$ \\
\bottomrule
\end{tabular}
\caption{Geometric diagnostics on the unmodified floating-point states.
$E_{\Y}$ is the squared distance per pixel to the training support; $A$
is the fraction retaining the nearest-support index of the initial
noisy state. The two reference maps use the same initial ensemble.
Boldface and underlining rank the five learned methods only.}
\label{tab:geometry}
\end{table}

%% file: appendices.tex
\section{Network and evaluation details}\label{supp:implementation}

\begin{figure}[!htbp]
\centering
\includegraphics[width=\linewidth]{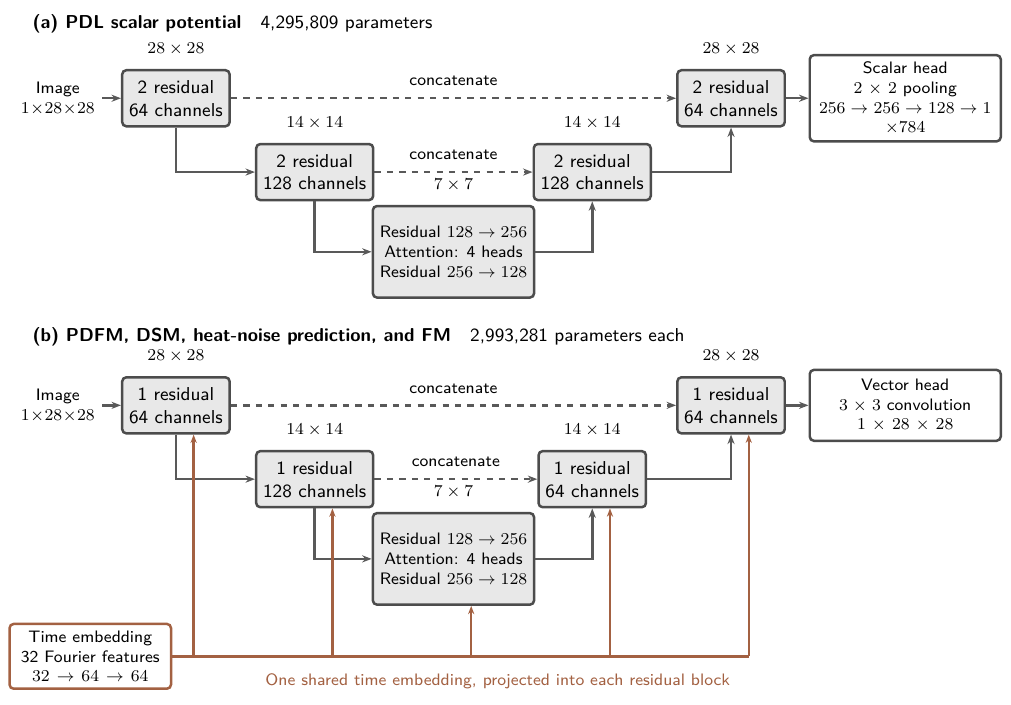}
\caption{Scalar and vector U-Nets. Dashed arrows concatenate encoder
features with decoder features; the brown arrows supply the shared time
embedding. PDL has ten residual blocks and $4{,}295{,}809$ parameters;
each vector model has six blocks and $2{,}993{,}281$ parameters.}
\label{supp:fig:architecture}
\end{figure}

Figure~\ref{supp:fig:architecture} shows the networks used in the MNIST
experiment. All operate on $1\times28\times28$ images and have base width
$64$. Their encoder, bottleneck, and decoder resolutions are $28$, $14$,
$7$, $14$, and $28$ in each spatial direction. Training objectives,
optimization, and the common generation protocol are specified in
Section~\ref{sec:datasets-targets}.

\FloatBarrier

\paragraph{Shared blocks.}
An initial $3\times3$ convolution produces $64$ channels. Each residual
block has two GroupNorm--SiLU--$3\times3$ convolution stages and an
identity or $1\times1$ shortcut. GroupNorm uses $32$ groups, and dropout
is zero. Downsampling and upsampling use channel-preserving $4\times4$
convolutions and transposed convolutions, respectively, with stride two
and padding one. The bottleneck consists of a $128\to256$ residual
block, four-head self-attention on the $49$ spatial positions, and a
$256\to128$ residual block.

\paragraph{Scalar and vector heads.}
PDL uses two residual blocks at each encoder and decoder level, with
encoder widths $64,128$ and decoder widths $128,64$. Before the final
concatenation, a $1\times1$ convolution reduces the upsampled width
from $128$ to $64$. The output features pass through
GroupNorm--SiLU--$3\times3$ convolution--GroupNorm--SiLU, then $2\times2$
adaptive average pooling and a SiLU MLP $256\to256\to128\to1$. Multiplication
by $d=784$ gives $\Phi_\theta$. The vector networks use one block per
level, encoder widths $64,128$, decoder widths $64,64$, and a final
GroupNorm--SiLU--$3\times3$ convolution to one image channel.

\paragraph{Time conditioning.}
For a positive input $c$, the vector networks form
$a=\log(\max\{c,10^{-8}\})/12$ and the $32$ features
$(\sin(\omega_j a),\cos(\omega_j a))_{j=0}^{15}$, with
$\omega_j=\exp(j\log1000/15)$. A $32\to64\to64$ SiLU MLP produces
the shared embedding; each residual block adds its own SiLU--linear
projection after the first convolution. The inputs are $c=e^{-\tau/2}$
for PDFM, $c=t$ for DSM and heat-noise prediction, and $c=e^r$ for FM,
where $r$ is its native interpolation time. PDL is autonomous.

\paragraph{Feature encoder.}
The frozen MNIST encoder has two $5\times5$ convolutions with $32$ and
$64$ channels, each followed by ReLU and $2\times2$ max pooling. A
fully connected ReLU layer maps $64\times7\times7$ features to a
$128$-dimensional embedding, followed by a ten-class linear classifier.
It was trained on all $60{,}000$ training images for five epochs with
cross-entropy and AdamW, learning rate $10^{-3}$, weight decay $10^{-4}$,
batch size $64$, and seed $3123$. Evaluation uses the original scale
of the frozen embedding.

\paragraph{Metric conventions.}
Metrics use clipped floating-point outputs before PNG conversion; the
raw states are retained for the geometric diagnostics. Pixel-nearest
reference ties are resolved by the first index in saved order.
SSIM uses zero padding, unit data range, and constants $0.01^2$ and
$0.03^2$. LPIPS uses AlexNet version $0.1$, bilinear resizing with
\texttt{align\_corners=False}, and input range $[-1,1]$. Feature FID
uses unbiased covariances with $10^{-6}I$ regularization and
double-precision moments and matrix square roots. KID uses the shared
subsets specified in the main paper and retains signed unbiased
estimates on their raw scale.

The accompanying arXiv ancillary archive supplies the exact code, parameter
files, reference-data and checkpoint identifiers, random seeds, and
complete numerical tables. The saved component grids and figure-layout
script reproduce all displayed panels.

\clearpage
\section{Generation trajectories}\label{supp:images}

\begin{figure}[!htbp]
\centering
\includegraphics[width=.96\linewidth]{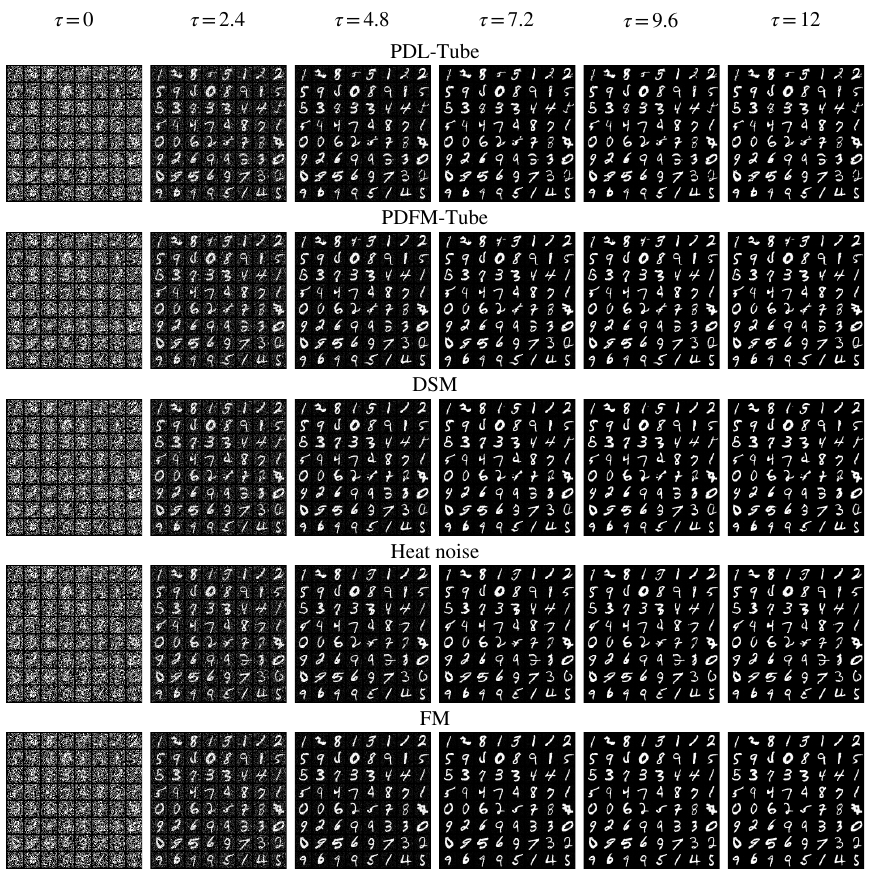}
\caption{Generation from the common noisy-data initialization.
Rows correspond to the five learned methods; columns show logarithmic
heat times $\tau=0,2.4,4.8,7.2,9.6,12$. The initial ensemble, integration
steps, and terminal heat time are the same across methods.}
\label{supp:fig:trajectories}
\end{figure}

Figure~\ref{supp:fig:trajectories} follows the same $64$ samples shown in
Figure~\ref{fig:comparison}. Each $8\times8$ grid uses the same sample
order, read from left to right and then from top to bottom. All images
are clipped to $[0,1]$ and use one grayscale range.

